# Double-Barreled Question Detection at Momentive

Peng Jiang [†]
Momentive
pjiang@momentive.ai

Krishna Sumanth Muppalla [†]
Momentive
krishnam@momentive.ai

Qing Wei [†]
Momentive
qwei@momentive.ai

Chidambara Natarajan Gopal
Momentive
cgopal@momentive.ai

Chun Wang
Momentive
chunw@momentive.ai

## ABSTRACT

Momentive is an agile experience management company, offering solutions in market research, customer experience, and enterprise feedback. The technology is gleaned from the billions of real responses to questions asked on the platform. When constructing a survey, it is important to make sure the survey questions are neutral to uncover unbiased opinions toward a subject. However, people may create biased questions even without intention. A double-barreled question (DBQ) is a common type of biased question that asks two aspects in one question. For example, a DBQ can be "Do you agree with the following statement: The food is yummy, and the service is great in this restaurant," with the answer options of "agree or disagree." This DBQ can confuse survey respondents because there are two parts in a question. If the survey taker's answer is "agree," there is no way to tell whether he/she agrees with just one part or all the parts. Thus, the responses are not accurate to reflect people's true thoughts. DBQs impact both the survey respondents and the survey owners. Momentive aims to detect DBQs and recommend survey creators to make a change towards gathering high-quality unbiased survey data.

Previous research work has suggested detecting DBQs by checking the existence of grammatical conjunction (i.e., and). While this is a simple rule-based approach, this method is error-prone because conjunctions can also exist in properly constructed questions. We present an end-to-end machine learning approach for DBQ classification in this work. As the percentage of DBQs is very low, we handled this imbalanced data using active learning. Then we compared many state-of-the-art embedding algorithms, including word2vec, FastText, BERT, XLNet, FB Infersent, and GoogleUSE to transform text data into vector representations used to train classification models. Furthermore, we proposed a model interpretation technique propagating the vector-level SHAP values to a SHAP value for each word in the question and answer text to interpret the model prediction better. We concluded that the word2vec subword embedding with maximum pooling is the optimal word embedding representation in terms of precision and running time in the offline experiments using the survey data at Momentive. The A/B test and production metrics indicate that this model brings a positive change to the business. To the best of our knowledge, this is the first machine learning framework for DBQ detection, and it successfully differentiates Momentive from the competitors. We hope our work sheds light on the machine learning approaches for bias question detection.

## 1 INTRODUCTION

Momentive is an agile experience management company delivering intuitive, people-centric solutions that help industry leaders quickly and confidently make important decisions, take actions, and achieve tangible results. Our technology collects billions of responses, with 25 million questions answered daily. Since our customers extract insights from these survey questions and response data, it is vital to ensure the data is of high-quality.

---

[†] The three authors contributed equally to this research

Creating unbiased questions is the first and essential step towards high-quality data.

There are many guidelines in survey research [1, 2] for constructing unbiased survey questions. However, survey creators may still easily make mistakes and create different types of biased questions with or without intentions. For example, people might develop questions like "How well would this app save time for you?" This is called a leading question that prompts a respondent towards providing an already-determined answer. In this case, the question indicates responses towards the assumption that the app helps save time. An improvement to this question could be "How might this app affect your efficiency, if at all?". Many other types of biased questions are summarized in [3, 4, 5]. This paper will focus on detecting a common type of biased question called double-barreled question.

Double-barreled question (DBQ) has two parts embedded in one question; this is where the word "double-barreled" comes in. For example, in the previous question, "Do you agree with the following statement: The food is yummy, and the service is great in this restaurant.". If the survey respondent's answer is "agree", he/she might agree that the food is yummy, or the service is great, or food is yummy and service is great. One could improve a DBQ by splitting it into two questions. People may ask DBQs with the intention of saving time to create surveys. However, DBQ is hard for respondents to understand which part of the question to answer. They might skip the questions, quit the survey, or provide answers unclear for survey creators to analyze. Either way, DBQ negatively impacts the respondent experience and the final survey results. Therefore, at Momentive we would like to detect DBQs when users create such questions so that they could make a quick fix to neutral questions, resulting in more unbiased responses.

Previous research work [7] has only suggested that DBQs can be detected by the existence of grammatical conjunctions such as "and", "along with", and "as well as". This approach is simple. However, there are several issues with this rule-based approach. First, conjunctions can exist in properly constructed questions. In the question "How much do you like the show 'Tom and Jerry'?", the conjunction "and" is a part of a show name, thus this question should not be detected as a DBQ. Second, not every DBQ contains conjunctions. For example, for the question "Do you agree that the app should improve its UI to deliver better customer experience?", an answer "yes" could refer to an agreement that the app should spend efforts on improving its UI, or the respondent agrees that improving UI is helpful for customer experience.

To address these challenges, we propose a deep learning framework to detect DBQs with explainability in prediction results. The model is trained using historical survey questions we have at Momentive. An in-house analysis indicates that the DBQs among the survey questions are small. Since imbalanced datasets usually result in models having poor predictive performance, we applied the active learning strategy [10, 11] to increase the chance of sampling DBQs to generate more balanced datasets for training. Next, we explore state-of-art word embedding algorithms such as Word2vec [14], FastText [15], BERT [16], XLNet [17] and Smooth Inverse Frequency (SIF) [12] with different pooling strategies [13]. We also evaluate several sentence embedding methods like FaceBook InferSent [18] and Google Universal Sentence Encoder (GoogleUSE) [19]. The embedding vectors are then fed into classification models to make predictions. To better explain the prediction results, we proposed a novel way to show the importance score of each word token to the prediction. Our approach is based on SHapley Additive exPlanations（SHAP）[20] by propagating the SHAP value of each dimension in the word embedding back to the word-level SHAP value. This provides a straightforward way to understand the importance of each word to the prediction, to improve the model in the next iteration.

Other than the aforementioned innovation technologies in our deep learning framework, we also differentiate ourselves to consider both question texts and answer options. Answer options provide important context into the judgment of DBQs. For example, if the question "are you pregnant or breastfeeding?" is paired with answer options of "yes" and "no", then it is a DBQ. However, if the answer options are "pregnant", "breastfeeding", "not pregnant", and "not breastfeeding", then it is a neutral question.

We conducted experiments on how we iteratively applied active learning to label question samples from different sources in our question bank. Experiments show that active learning successfully increases the ratio of DBQs in the training data. Comparing the embedding algorithms with the rule-based approach shows that the word2vec subword with maximum pooling is the optimal word embedding representation in terms of precision and running time. In the A/B test, this algorithm was applied in the test group to flag DBQs for survey creators and recommend them to make a change. There is no DBQ prediction in the control group. The A/B test results indicate this model lifts the business metrics; therefore, the solution was deployed in production to enhance the capabilities of our survey creation tool SurveyMonkey Genius [6]. Momentive builds this tool to help our customers make better and faster surveys highlighting potential biases and errors in survey questions and estimating survey completion time and completion rate.

To the best of our knowledge, this paper is the first to propose an end-to-end deep learning framework to detect a common biased question type of DBQ, and our framework successfully helps Momentive achieve a competitive advantage in the survey creation domain. We addressed the imbalance challenges in the in-house data, evaluated and identified the best embedding algorithm, and proposed a novel approach to explain model predictions at the word level. The A/B test results indicate that our DBQ detection model could boost business metrics. The integration architecture is provided to illustrate how the model is deployed in our platform. Our logged metrics show that a good percentage of our customers will change their original questions after observing the questions flagged by our model.

## 2  PROBLEM STATEMENT

Momentive provides several phases to guide our customers from creating surveys to analyzing their survey results for invaluable insights. The phases include creating surveys, previewing surveys, collecting responses, and analyzing results. DBQ detection and recommendations are related to the first two phases. The first phase is where customers come to our platform and design surveys. They will create survey questions to gather insights for their purpose, as shown in Figure 1. In the next phase, we provide the in-house tool, SurveyMonkey Genius, for survey creators to preview their surveys and make necessary changes to enhance the survey quality. Our tool offers insights such as predicted survey completion rate, completion time, and recommendations for correcting potential biases and errors in survey questions. This is a critical step impacting respondents' experience in answering survey questions and survey creators for extracting unbiased insights.

Our model is integrated into the SurveyMonkey Genius tool to flag DBQs and surface recommendations. Figure 2 shows that our UI highlights the first question as a DBQ based on model predictions, explains the definition of DBQ, and recommends a simple change of splitting a DBQ into two questions to make it neutral. Customers may click the "applied to question 1" link to manually update that question. We log the click rate of the link to track user engagement.

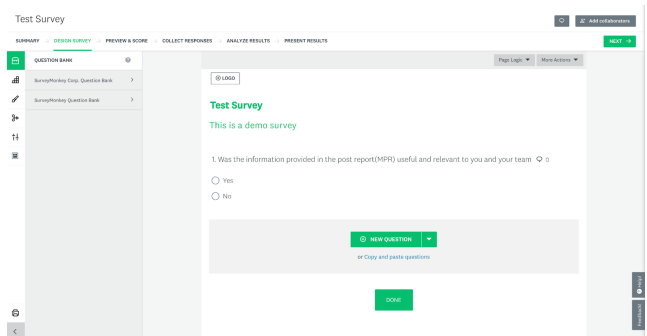

**Figure 1: Survey Creation Phase**

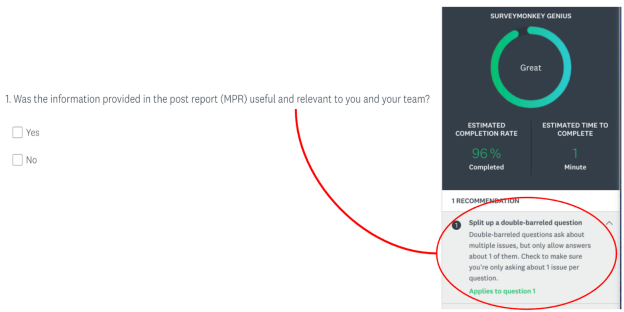

**Figure 2: Survey Preview Phase**

## 3 FRAMEWORK

In this section, we describe our adoption of active learning with a committee-based sampling strategy to handle imbalanced data, the exploration of several state-of-art embedding algorithms with different pooling strategies, and our proposed way of generating a SHAP value for each word/phrase in the questions to interpret model predictions. In addition, we illustrate the deployment architecture of the model into our platform.

### 3.1 Active Learning

DBQs are a small proportion of the questions we process based on our in-house analysis. If we employ the random sampling strategy to generate training data from these questions, the ratio of DBQs will remain low. Research has shown that training machine learning models with an imbalanced dataset can lead to biased prediction and decrease model performance [8, 9]. Meanwhile, the cost of labeling DBQs is expensive. As mentioned in the introduction, DBQs do not always contain conjunctions, which makes the labeling efforts costly.

Active learning [10] is well known for reducing labeling efforts. Instead of randomly choosing a sample from the unlabelled set for humans to label, active learning applies querying strategies to pick the most informative sample to query its label. In this work, we apply the vote entropy implemented in modAL [21], an active learning framework for Python3, to define the informativeness of each sample. And we use three classifiers as a committee: Random Forest, Logistic Regression, and Gradient Boosted Regression Trees to calculate the vote entropy of each sample. A sample is considered as most informative if there is maximum disagreement in the prediction votes of the committee. The vote entropy formula for measuring the disagreement is listed below:

$$H(x) = -\sum_{k=1}^{2} q_k * \log(q_k)$$

where $x$ refers to a sample, $k$ is a particular class in a classification problem, and $q_k$ is the number of votes for $k$-th class divided by the total number of classifiers. In our binary classification problem, $k$ ranges from 1 to 2. If two classifiers predict a sample as DBQ, and one classifier predicts it non-DBQ, then $q_1$ and $q_2$ will be $\frac{2}{3}$ and $\frac{1}{3}$ respectively.

We describe in the experiment section how we apply this strategy iteratively to query question samples from different sources in our question bank. The experiment shows that active learning successfully increases the ratio of DBQs in the training data.

### 3.2 Embedding with Pooling Strategies

This section introduces the application of state-of-the-art word embedding and sentence embedding models into our problem. Figure 3 shows how we apply the word2vec (w2v) method to generate a sentence-level embedding for a question and an answer-option-level embedding for its answer options, then concatenate both to get the final embedding for classification.

Different pooling strategies are explored in this study. For instance, we have $\{v_1, v_2, ..., v_L\}$ denote the token embeddings, where $L$ is the number of tokens and $v_i \in \mathbb{R}^K$, then we can define the output of mean-pooling as $z = \frac{1}{L}\sum_{i=1}^{L} v_i$, sum-pooling as $z = \sum_{i=1}^{L} v_i$, max-pooling as $z = \max\{v_1, v_2, ..., v_L\}$ and min-pooling as $z = $ $\min\{v_1, v_2, ..., v_L\}$, where max-pooling and min-pooling are defined on element-wise comparisons among the embedding vectors. The left side of figure 3 shows how we use max-pooling to convert embeddings of words in the question to an embedding vector of the whole question. The right side of figure 3 shows how we use mean-pooling to convert embeddings of phrases in each answer option to an embedding vector of all answer options. Specifically, we first average the individual embedding for "strong" and "agree" to get the representation of the answer options "strongly agree", then we average the representations of 3 answer options to get the representation of all answer options together. The bottom of figure 3 shows a simple concatenation to get the final representation for this sample pair of question and answer options. The numerical representations of all samples are then used as features in classification models.

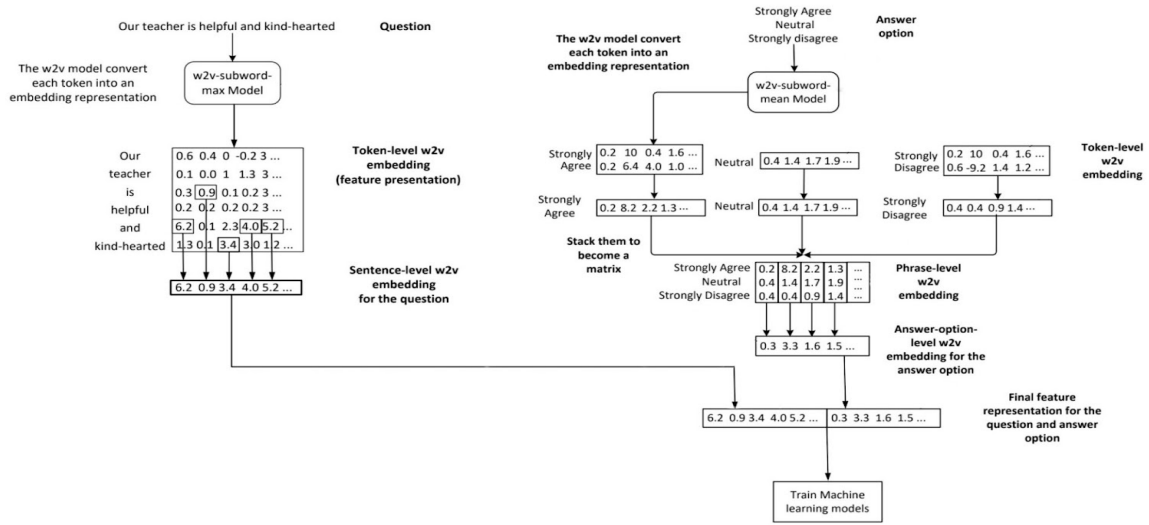

**Figure 3: Embeddings with Pool Strategies for a Sample Question**

We later compared the combinations of different embedding algorithms with various pooling strategies using the in-house data. The optimal way of combining embedding algorithms with pooling strategies is identified for DBQ detection.

### 3.3 Model Explainability

To intuitively understand model predictions, we propose a novel interpretation technique by propagating the vector-level SHAP value back to the word-level SHAP representation leveraging the pooling strategies. The output of the technique is a numerical value that can represent the importance of every word token to the DBQ prediction.

Figure 4 uses the previous sample question and answers options in Figure 3 to show how we make use of different pooling strategies to obtain the word-level SHAP value. The vector on the top is the SHAP value for each dimension in the embeddings of a pair of the question and answer options. For example, the SHAP value of 4 is for the first dimension in the question embedding, and 6 is the SHAP value of the first dimension in the answer option embedding. The token-level and phrase-level w2v embedding matrices refer to the same ones shown in Figure 3. We apply max-pooling and mean-pooling respectively to these two matrices to generate embeddings of the question and answer options. The projected SHAP matrices are generated column by column using the vector-level SHAP values and the corresponding columns in the embedding matrices, indicated by the red and blue arrows. For example, the SHAP value 4 is transformed to a column vector of $[0, 0, 0, 0, 4, 0]^T$ in the project SHAP matrix. As when applying the max-pooling to the first column of the w2v embedding matrix to generate the first word embedding value for the question, it will pick the 5th word "and". Therefore, we give all credit to "and". On the right-hand side of Figure 4, the SHAP value of 6 is shared based on the phrase-level embedding matrix to generate the values in the projected SHAP matrix. For example, "strongly agree" has a SHAP value of 1.2 in the first dimension because this phrase contributes 20% (i.e., $0.2/(0.2 + 0.4 + 0.4)$) in the embeddings, thus it is assigned 20% of the SHAP value 6, which is 1.2. The update formula is as follows:

$$SHAP_{Matrix_{i,j}} = SHAP_{Original_j} * \frac{EmbeddingMatrix_{i,j}}{\sum_{k=1}^{\#of\ tokens} EmbeddingMatrix_{k,j}}$$

The projected SHAP matrix has the same dimension as the embedding matrix. Each word or phrase in the SHAP matrix is represented by a row of SHAP values. The final SHAP value for a word or phrase is an aggregation of the values in that row. One could use summation, average them, or take the maximum of all values in the row. This strategy provides a straightforward way to visualize the contributions of words and phrases in questions and answer options to the model predictions.

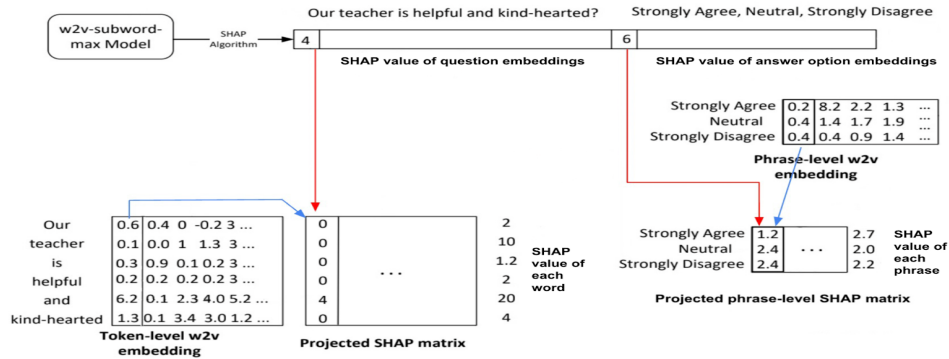

Figure 4: Proposed Interpretation Technique for Token-Level SHAP Value

### 3.4 Integration Architecture

As shown in Figure 2, DBQ detection is deployed in the survey preview stage to recommend our customers to make a change. Here we describe the integration details in Figure 5. The DBQ model is deployed as a containerized web service based on python pyramid. When the docker container is spined up, the service will preload the model artifact from AWS S3, indicated as the step (0). Next, we will describe the flow from steps (1) to (6).

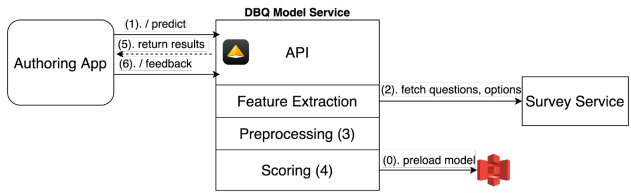

Figure 5: Integration Architecture

(1). When customers finish creating surveys and go to the Preview page, the Authoring App (our survey editing app) will send a predict request to DBQ Model Service with survey_id in the payload. This is a synchronous call. The frontend shows a loading spinner and waits for the results to display.
(2). DBQ Model Service will then call SurveyService (our web service which manages survey forms) to fetch the questions along with their answer options.
(3). Then the raw features will be fed to the preprocessing pipeline to generate vectorized features.
(4). Once vectorized features are ready, the model will be called to detect DBQs.
(5). The results will be returned to the client (Authoring App) and displayed to end-users.
(6). If customers take our recommendation and correct DBQs, the feedback will be sent to ML service. Thus, we could use the feedback to evaluate model performance and improve the model in future iterations.

## 4 EMPIRICAL RESULTS

This section shows the training and evaluation of the DBQ model using Momentive's in-house data. Specifically, it illustrates our adoption of active learning to generate training and test datasets, the comparison of several baseline models, the evaluation of different embedding algorithms with pooling strategies, the explanation of model predictions on real customer questions, and the business impact of our solution.

### 4.1 Active Learning for Data Sampling

Momentive processes a huge amount of survey questions and answers in multiple languages. To train the deep learning model, we retrieve 203,411 surveys with 1,481,853 questions from the survey database. Moreover, we focus on English surveys between mid-2017 to mid-2018, and only questions limited to 200 words are picked. Each sample in the training is defined as an input pair of question text and its corresponding answer option text.

To apply active learning, we need a seed, i.e., a small set of labeled samples. Data is sampled from different sources: questions that contain conjunctions and those not containing conjunctions. While it is hard to provide guidelines to label DBQs containing no conjunctions, we consult with our survey research team to give labelling instructions for the other scenario. We define a list of seven coordinating conjunctions: "for", "and", "nor", "but", "or", "yet", and "so", and five synonyms of "and": "along with", "also", "as well as", "together with", "including". A question is labeled as DBQ if it contains at least one conjunction, and the words/phrases before and after the conjunction(s) have no clear association, causation, or entailment relationships.

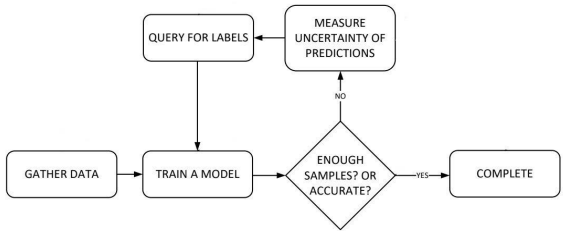

**Figure 6: Active Learning Loop for Data Sampling**

Figure 6 provided by modAL [22], an active learning framework for Python3, shows that the active learning workflow usually repeats the sampling process until enough samples are collected or the model performance meets the need. We repeat the loop twice to collect 2000 and 6000 question samples respectively in each iteration. The specific steps are listed below:

1. Retrieve 5,000 surveys each from two different sources: surveys having at least one question with predefined conjunctions, and surveys having no question containing predefined conjunctions.
2. Use active learning associated with committee-based sampling strategy to query 1,000 questions from each data source.
3. Use the labeled data as ground truth and rerun Active Learning, then label an additional 2,000 questions from each data source.

We get 6000 question samples with active learning workflow, among which 5,000 questions are labelled for training. The remaining 1,000 questions are for testing. We refer it Test set 1. In addition, we sampled another 1,500 questions with random sampling strategy and refer to this as Test set 2. Figure 7 illustrates the sampling process. And Table 1 shows the statistics of the two test sets. Compared with random sampling, active learning increases the chance of sampling DBQs from 5.13% to 21.7%.

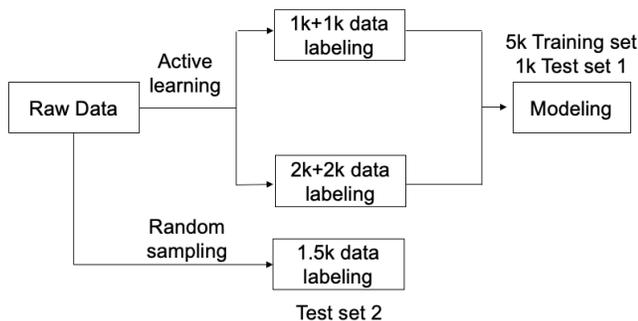

**Figure 7: Data Sampling using Active Learning and Random Sampling**

|  | Test set 1 | Test set 2 |
|---|---|---|
| Sample Size | 1000 | 1500 |
| DBQs | 217 | 77 |
| DBQ ratio | 21.7% | 5.13% |

**Table 1: Statistics of Test Sets**

## 4.2 Performance Evaluation

In the experiment we compared state-of-art word-level embedding methods including Word2vec, FastText, BERT, XLNet and Smooth Inverse Frequency (SIF) associated with mean-pooling, max-pooling, min-pooling and sum-pooling strategies. Also, we test the sentence embedding methods FaceBook InferSent and Google Universal Sentence Encoder (GoogleUSE). The embedding vectors are used in traditional classification models for DBQ classification. Our experiments use Random Forest as the classification model.

We use three popular classification metrics, precision, accuracy, and AUC-ROC value, to compare model predictions. Precision is chosen for model selection decisions in our problem. That is because when we flag neutral questions as DBQs and suggest our customers make a change towards these questions, it would be confusing to them and affect the customer trust.

### 4.2.1 Baselines

We first evaluate several simple baselines and choose the most appropriate one for further comparison with the deep learning framework. The first two baselines are based on the simple rule-based approach of checking the conjunctions such as "and".

**Approach 1**: In a question text, there is at least one conjunction.

**Approach 2**: In a question text, there is at least one conjunction. Another constraint is that the text has only one sentence.

However, as described in the Introduction section, conjunctions can exist in properly constructed questions such as "How much do you like the show 'Tom and Jerry'?". Thus, we propose to first apply the NLP chunking module in spaCY to remove phrases like "Tom and Jerry" before checking the conjunctions. The third baseline is described as follows:

**Approach 3**: First remove noun phrases using NLP chunking, then apply approach 2.

Since all these baseline approaches do not require training data, we use the whole set of 6,000 labeled samples chosen by active learning to compare their performance. Table 2 shows the evaluation results. To our surprise, the chunking algorithm does not work as expected to help boost precision. This indicates that we need a more sophisticated solution for DBQ detection. Approach 2 is chosen as the final baseline for later comparison, as it has the highest precision.

|  | Precision | Recall |
|---|---|---|
| Approach 1 | 0.5000 | **0.886** |
| Approach 2 | **0.538** | 0.814 |
| Approach 3 | 0.472 | 0.835 |

Table 2: Performance Comparison of Baselines

| Test Dataset | Methods | Precision | Accuracy | AUC-ROC |
|---|---|---|---|---|
| Test set 1 | w2v-subword-max | **0.761** | 0.841 | 0.902 |
| Test set 1 | GoogleUSE | 0.717 | **0.865** | **0.917** |
| Test set 1 | Baseline | 0.542 | 0.852 | NA |
| Test set 2 | w2v-subword-max | **0.583** | **0.955** | 0.889 |
| Test set 2 | GoogleUSE | 0.472 | 0.951 | **0.894** |
| Test set 2 | Baseline | 0.179 | 0.846 | NA |
| Test set 1 + Test set 2 | w2v-subword-max | **0.697** | 0.905 | 0.895 |
| Test set 1 + Test set 2 | GoogleUSE | 0.614 | **0.914** | 0.914 |
| Test set 1 + Test set 2 | Baseline | 0.367 | 0.847 | NA |

Table 3: Performance Comparison of Deep Learning Approaches with the Baseline

### 4.2.2 Offline Evaluation of Embedding Algorithms

We compared different word and sentence embedding algorithms and identified 128-dimensional word2vec subword with maximum pooling (referred as w2v-subword-max) is the most promising word embedding algorithm, and 512-dimensional GoogleUSE performs best among all sentence embedding algorithms. Here we omit the performance comparison of all embedding algorithms. Instead, we compare the performance of the two embedding algorithms with the best identified baseline (approach 2) on the two test sets described in Section 4.1. As a reminder, test set 1 is sampled with active learning strategy, and test set 2 is generated with random sampling.

The performance results shown in Table 3 indicates several conclusions: First, the accuracy of all three methods does not differ much when evaluated on different test sets. The same conclusion holds for AUC scores of the two embedding algorithms. Second, for our interested metric precision, both embedding algorithms performed much better than the baseline in all 3 test cases. And w2v-subword-max has the highest precision. Third, the precision of all three methods drops when testing on test set 2 compared with test set 1. And the baseline model's precision drops a lot more when comparing with both embedding algorithms. Remember that test set 1 has a similar distribution with the training data, and test set 2 has a much less ratio of DBQs. This explains the performance drop in embedding methods due to the difference of data distribution between training and test. And the precision decrease of the baseline in test set 2 could be due to the fact that the ratio of non-DBQs is very high (~95%), so the chance of predicting a non-DBQ to DBQ increases, therefore affecting the precision.

Other than comparing the performance metrics, we would like to share that our model could detect DBQs with no conjunctions. For example, our model detect this DBQ from our question bank, "I could openly express my opinion about organization/processes.".

Next, we show the time comparison of the two embedding algorithms. As in production we will run the model for all questions in a survey, the comparison is performed at the survey level. Table 4 shows the running time per survey on sampling surveys with two different lengths. Our in-house analysis indicates the average survey length is between 10-20 questions, so we sample surveys having 20 questions in our first set. The other set covers the extreme case in which each survey includes 200 questions. The running time includes two parts: time for embedding algorithms to generate vector representations, and time for classification models to make predictions. We notice that the classification time does not differ much between embedding algorithms on different survey sets, while w2v-subword-max beats GoolgeUSE as it needs less time for embedding representations regardless of the survey length.

| Surveys | Methods | Embedding | Classification |
|---|---|---|---|
| Small surveys with 20 questions | w2v-subword-max | **0.052s** | 0.102s |
| | GoogleUSE | 0.088s | 0.153s |
| Large surveys with 200 questions | w2v-subword-max | **0.549s** | 0.103s |
| | GoogleUSE | 0.633s | 0.103s |

**Table 4: Time Comparison on Surveys**

### 4.2.3 Experiments with Additional Features

Other than using embedding to get semantic features, we also experiment using additional handcraft features. Based on domain experts' feedback, we identify a list of about 50 features including number of words in a question, number of noun phrases, the presence of part-of-speech tags and punctuations to reflect the sentence structure.

Table 5 lists the comparison of only semantic features generated by embedding algorithms and with additional handcraft features. In this experiment, we use w2v-subword-max as the embedding algorithm, as it performs better than GoogleUSE in precision and running time. Table 5 shows that adding additional features could improve AUC values across all test datasets for about 3%, increase accuracy for about 2% for some test datasets, but decrease precision by around 6% to 17% across various test cases. Combining the observations in section 4.2.2 and 4.2.3, we choose w2v-subword-max as the embedding method to use with random forest for DBQ classification.

| Test Dataset | Features | Precision | Accuracy | AUC-ROC |
|---|---|---|---|---|
| Test set 1 | only semantic features | **0.761** | 0.841 | 0.902 |
| | additional features | 0.709 | **0.865** | **0.929** |
| Test set 2 | only semantic features | **0.583** | 0.955 | 0.889 |
| | additional features | 0.412 | 0.946 | **0.911** |
| Test set 1 and 2 | only semantic features | **0.697** | 0.905 | 0.895 |
| | additional features | 0.591 | **0.911** | **0.920** |

**Table 5: Performance Comparison with Handcraft Features**

### 4.2.4 Model Explainability for Question Pairs

We now show the visualization of our proposed explainability mechanism for the predicted questions. provide SHAP values of words/phrases. Figure 8 shows the visualization of model predictions on question pairs, and both samples are true positives. In the first question pair of "Do you feel that the central office administrators listen to your concerns and attempts to remedy those concerns?" and its associated answer options "Answer option: Yes, No, Comment", the conjunction word "and" has the largest SHAP value among all tokens. This is in line with how rule-based approach uses conjunctions for DBQ classification. In the second question pair of "The information process was simple and effective." and its associated answer options "Strongly agree, agree, neutral, disagree, strongly disagree, N/A", the adjustive "simple" has the maximum SHAP value among all words in the question, and another adjustive "effective" also ranks relatively high of being the 4[th] largest SHAP value. The two adjustives surrounding the keyword "and" indicates that the sentence structure plays an important role in DBQ detection. Meanwhile the SHAP values in the answer options do not differ much.

Question: Do you feel that the central office administrators listen to your concerns and attempts to remedy those concerns?

Answer option: Yes, No, Comment

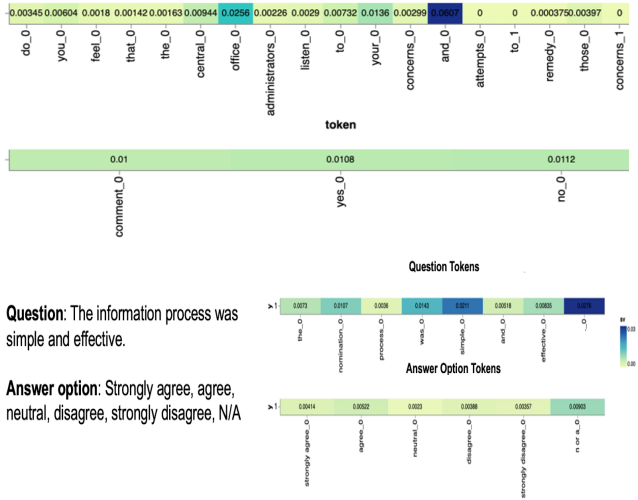

**Figure 8: Word/Phrase Level Explainability for Two Examples**

Compared with the normal SHAP values for each embedding dimension, the word/phrase level interpretability provides a straightforward way to understand model predictions and therefore help improve model performance in future iterations.

#### 4.2.5 Online Evaluation

Our DBQ model was in production in early 2020 after the A/B test. Due to confidentiality, we will share the conclusions omitting details. The A/B test was done from May 22 to June 25, 2020 with users split equally between control and test groups. The control group does not have any DBQ recommendation, while the test group is exposed to this feature. The result showed an increase in the deployment rate by 4.5% with the DBQ feature. Here deployment rate is an important business metric to measure survey success at Momentive. It is defined as the percentage of surveys receiving at least 5 responses among all created surveys. This is in line with our expectation, as correcting DBQs would help remove the confusion put into survey respondents, therefore reducing the chance of people skipping questions in surveys.

The DBQ feature was deployed in production after observing the A/B test result. Our logging data indicates that the adoption rate for the DBQ feature is high. The adoption rate is ratio of users who applied the DBQ recommendations among all users.

## 5 CONCLUSION AND FUTURE DIRECTION

We proposed an end-to-end deep learning framework for detecting DBQs. Our framework includes the adoption of active learning for sampling higher ratio of DBQs, the exploration of various embedding algorithms, the comparison of semantic features with handcrafted features, and the novel approach to provide token-level SHAP value to better understand model predictions. The A/B test shows an increase of survey deployment rate with the DBQ model. Many customers reply on model predictions to help them mitigate the biases, therefore benefiting survey respondents and survey owners. In future we plan to explore how to better make use of the explainability visualization to improve model performance.

## ACKNOWLEDGEMENTS
We would like to thank King Chung Ho, Jing Huang, Braden Dong, Melanie Lei, Geet Chopra, Meghan Horvath, Jarmila Henn for their helpful comments and suggestions from legal, external communication, product, and methodology perspectives.

## ONLINE RESOURCES
We will consider individual requests to access the source code.

## REFERENCES

[1] Weisberg, H. F., Krosnick, J. A., Bowen, B. D., & Weisberg, H. F. 1996. An introduction to survey research, polling, and data analysis.
[2] David De Vaus. 2002. Surveys In Social Research. Fifth edition. Routledge, 97-99
[3] Thyer, Bruce. The handbook of social work research methods. Sage Publications, 2009.
[4] SurveyMonkey. 2020. 5 common survey question mistakes that'll ruin your data.https://www.surveymonkey.com/mp/5-common-survey-mistakes-ruin-your-data/
[5] Gail M. Sullivan, MD, MPH and Anthony R.Artino Jr, PHD. 2017. How to Create a Bad Survey Instrument. In Journal of Graduate Medical Education, August 2017.
[6] SurveyMonkey. 2020. SurveyMonkey Genius. https://www.surveymonkey.com/mp/surveymonkey-genius/
[7] Earl R. Babbie, Lucia Benaquisto, Fundamentals of Social Research, Cengage Learning, 2009, Google Print, p.251.
[8] Q. Gu, Z. Cai, L. Zhu, and B. Huang, "Data mining on imbalanced data sets," in 2008 International Conference on Advanced Computer Theory and Engineering, 2008, pp. 1020–1024.
[9] Q. Wei and R. L. Dunbrack Jr. The role of balanced training and testing data sets for binary classifiers in bioinformatics. in PLoS One, vol. 8, no. 7, 2013.
[10] B. Settles. 2009. Active learning literature survey. In Computer Sciences Technical Report 1648, University of Wisconsin-Madison
[11] Ertekin, Seyda, et al. "Learning on the border: active learning in imbalanced data classification." Proceedings of the sixteenth ACM conference on Conference on information and knowledge management. 2007.
[12] Sanjeev Arora, Yingyu Liang, Tengyu Ma. A Simple but Tough-to-Beat Baseline for Sentence Embeddings, in ICLR 2017 Conference, 04 Nov 2016.
[13] Dinghan Shen, Guoyin Wang, Wenlin Wang, Martin Renqiang Min, Qinliang Su, Yizhe Zhang, Chunyuan Li, Ricardo Henao, Lawrence Carin. Baseline Needs More Love: Word-Embedding-Based Models and Associated Pooling Mechanisms. In arXiv: 1805.09843v1, 24 May 2018.
[14] T. Mikolov, I. Sutskever, K. Chen, G. S. Corrado, and J. Dean. Distributed representations of words and phrases and their compositionality. In Advances in neural information processing systems, 2013, pp. 3111–3119.
[15] P. Bojanowski, E. Grave, A. Joulin, and T. Mikolov. Enriching Word Vectors with Subword Information. In Trans. Assoc. Comput. Linguist., vol. 5, pp. 135–146, 2017.
[16] J. Devlin, M.-W. Chang, K. Lee, and K. Toutanova. Bert: Pre-training of deep bidirectional transformers for language understanding. In arXiv Prepr. arXiv1810.04805, 2018.
[17] Zhilin Yang, Zihang Dai, Yiming Yang, Jaime Carbonell, Ruslan Salakhutdinov, Quoc V. Le. 2020. XLNet: Generalized Autoregressive Pretraining for Language Understanding. In arXiv:1906.08237, 2020.
[18] A. Conneau, D. Kiela, H. Schwenk, L. Barrault, and A. Bordes. 2017. Supervised Learning of Universal Sentence Representations from Natural Language Inference Data. May 2017.
[19] D. Cer et al. 2018. Universal Sentence Encoder. Mar,2018.
[20] S. M. Lundberg and S.-I. Lee. 2017. A unified approach to interpreting model predictions. In Advances in neural information processing systems, pp. 4765–4774.
[21] Disagreement sampling in modAL. https://modal-python.readthedocs.io/en/latest/content/query_strategies/Disagreement-sampling.html
[22] modAL, an active learning framework for Python3. https://modal-python.readthedocs.io/en/latest/content/overview/modAL-in-a-nutshell.html